\documentclass[10pt,twocolumn,letterpaper]{article}

\usepackage{iccv}
\usepackage{graphicx}
\usepackage{multirow}
\usepackage{array}
\usepackage[export]{adjustbox}
\usepackage[most]{tcolorbox}
\usepackage{lipsum}
\usepackage{parskip}

\definecolor{iccvblue}{rgb}{0.21,0.49,0.74}
\usepackage[pagebackref,breaklinks,colorlinks,allcolors=iccvblue]{hyperref}
\usepackage{float}

\def\paperID{15} 
\def\confName{ICCV}
\def\confYear{2025}

\title{\textit{GenAI Confessions:} \\ Black-box Membership Inference for Generative Image Models}  

\author{Matyas Bohacek \\
Stanford University \\
{\tt \small maty@stanford.edu}
\and
Hany Farid \\
University of California, Berkeley \\
{\tt\small hfarid@berkeley.edu}
}

\begin{document}
\twocolumn[{%
\renewcommand\twocolumn[1][]{#1}%
\maketitle

\centering

    \fbox{

        \includegraphics[width=0.975\textwidth]{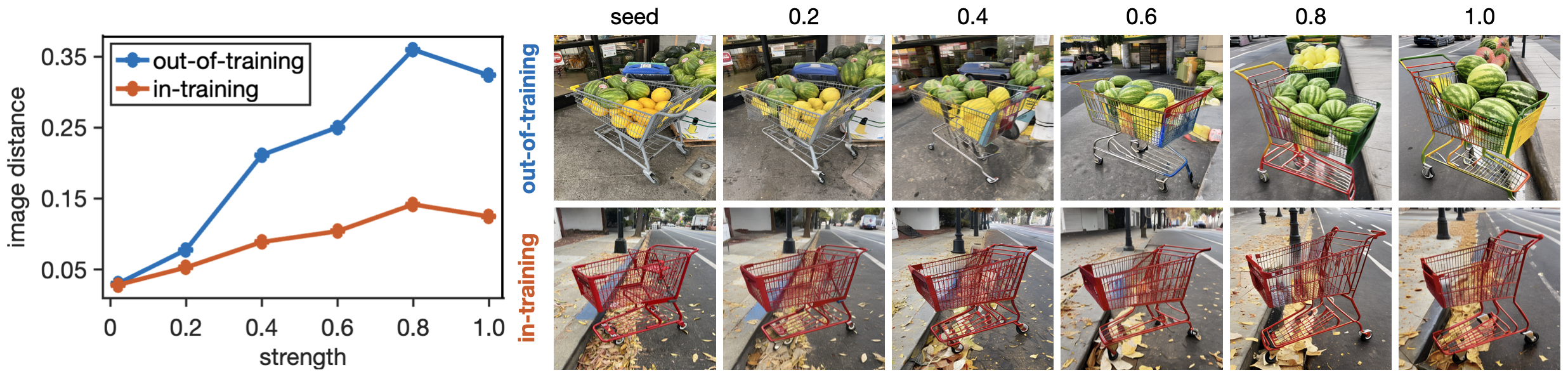}
        }

\captionof{figure}{\textbf{Overview of Our Membership Inference Method.} Images are generated using an image-to-image pipeline, seeded with an in-training image (bottom) and an out-of-training image (top), across increasing strengths. A larger strength corresponds to reduced emphasis on the seed image. The plot on the left shows the similarity between each generated image and its corresponding seed image (the DreamSim image similarity metric), where smaller values indicate greater similarity. The in-training seed consistently yields more similar outputs than the out-of-training seed. Our membership inference method exploits this phenomenon. \vspace{2em}}

\label{fig:teaser}

}]

\maketitle

\begin{abstract}
From a simple text prompt, generative-AI image models can create stunningly realistic and creative images bounded, it seems, by only our imagination. These models have achieved this remarkable feat thanks, in part, to the ingestion of billions of images collected from nearly every corner of the internet. Many creators have understandably expressed concern over how their intellectual property has been ingested without their permission or a mechanism to opt out of training. As a result, questions of fair use and copyright infringement have quickly emerged. We describe a method that allows us to determine if a model was trained on a specific image or set of images. This method is computationally efficient and assumes no explicit knowledge of the model architecture or weights (so-called black-box membership inference). We anticipate that this method will be crucial for auditing existing models and, looking ahead, ensuring the fairer development and deployment of generative AI models.
\end{abstract}

\vspace{-1em}
\begin{minipage}{0.48\textwidth}  
\begin{tcolorbox}[colback=gray!10, colframe=gray!20, boxrule=0pt, arc=2pt, left=6pt, right=6pt, top=4pt, bottom=4pt]
\small \textbf{Project Page:}~~~~~~~~~~\href{https://genai-confessions.github.io}{\texttt{genai-confessions.github.io}} \\
\small \textbf{Data:}~~~~~~~~~~~~~~~~~~~~\href{https://huggingface.co/datasets/faridlab/stroll}{\texttt{hf.co/datasets/faridlab/stroll}}
\end{tcolorbox}
\end{minipage}

\section{Introduction}

Most agree that -- despite occasional hallucinations, extra fingers and toes, and gravity-defying motion -- AI-powered systems are now capable of creating human-like prose, image, and video from a simple prompt. Most also agree that these systems, in the form of large-language~\cite{minaee2024large}, image~\cite{raut2024generative}, and video~\cite{melnik2024video} models, are only possible thanks to the ingestion of massive amounts of human-generated content. Here, however, is where disagreements begin~\cite{epstein2023art}. 

As the courts -- and the court of public opinion -- adjudicate these matters in the coming years, the question of whether a model was trained on a specific dataset will be critical. This so-called question of {\em membership inference} is challenging for a number of reasons. First, training datasets are massive and often created through large-scale web scraping without careful record keeping~\cite{whang2023data}. Second, once trained, the models are opaque, making a post-hoc inference challenging. And third, given the competitive landscape and lack of clear laws, there is currently little incentive for rule following, with even some former tech CEOs encouraging young entrepreneurs to steal intellectual property and later hire lawyers to ``clean up the mess''~\cite{vergeExGoogle24}.

The study of membership inference emerged as deep-learning applications started to be trained on large datasets~\cite{hu2022membership}. Early work focused on membership inference targeting classification (as compared to generative) models~\cite{shokri2017membership}. More recently, attention has turned to membership inference for generative models, including large language models (LLMs)~\cite{mattern2023membership,duan2024membership}, generative adversarial networks (GANs)~\cite{hayes2017logan,hilprecht2019monte}, and diffusion-based text-to-image models~\cite{wu2022membership,carlini2023extracting,zhai2024membership,li2024unveiling,zhang2024generated}. 

Diffusion-based models are the leading contenders in producing photorealistic images and video and hence the focus of our effort. Previous membership inference methods for these models either assume full or partial knowledge of the generative model architecture and trained weights (white-box or gray-box), are applicable to only one model architecture, require massive computing power to operationalize, or only work for analyzing membership for an entire dataset, as compared to a single image.

By contrast, our membership inference assumes no explicit knowledge of the model details (black-box), generalizes to different model architectures, is computationally efficient, and can operate on both a dataset of images as well as a single image.

The contributions of this paper can be summarized as follows:

\begin{itemize}
    \item A computationally efficient and easy to implement, black-box membership inference method for generative-AI image models (Figure~\ref{fig:teaser});
    \item A dataset (\textbf{STROLL}) of semantically matched image pairs for evaluating membership inference;
    \item Empirical analysis of membership inference and memorization across different model architectures.
\end{itemize}



\section{Methods}
\label{sec:methods}

\subsection{Data}

We compiled three datasets consisting of images generated by Stable Diffusion (v1.4, v2.1, v3.0), Midjourney (v6), and DALL-E (v2). As described below, these datasets consist of paired in-training and out-of-training images used to evaluate our membership inference technique. Each pair of images is constructed to be semantically similar in terms of content so as to ensure that any observed differences between in-training and out-of-training seed images is not due to semantic differences.

\begin{figure}[p]
    \centering
    \vspace{0.2cm}
    \fbox{
    \begin{tabular}{p{0.02\textwidth}p{0.11\textwidth}@{\hspace{8pt}}p{0.11\textwidth}@{\hspace{8pt}}p{0.11\textwidth}}
        \raisebox{0.15\height}{\rotatebox{90}{\shortstack{\small \textbf{in-training} \\ {\scriptsize ~}}}} & 
        \includegraphics[width=0.11\textwidth]{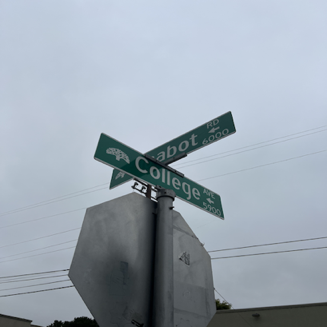} & 
        \includegraphics[width=0.11\textwidth]{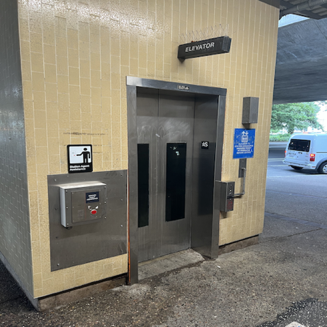} & 
        \includegraphics[width=0.11\textwidth]{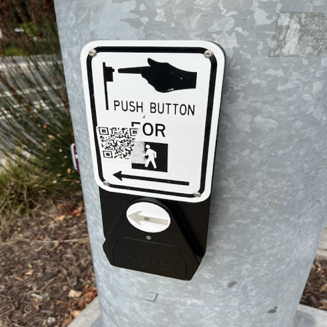} \\
        \raisebox{0\height}{\rotatebox{90}{\shortstack{\small \textbf{out-of-training} \\ {\scriptsize (DALL-E)}}}} & 
        \includegraphics[width=0.11\textwidth]{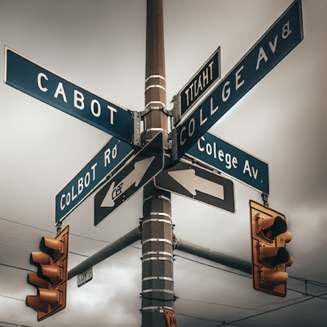}  &
        \includegraphics[width=0.11\textwidth]{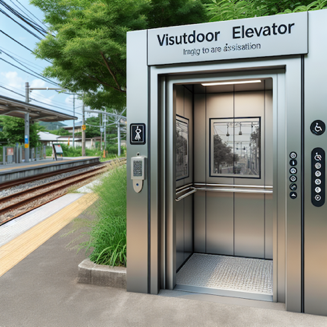} &
        \includegraphics[width=0.11\textwidth]{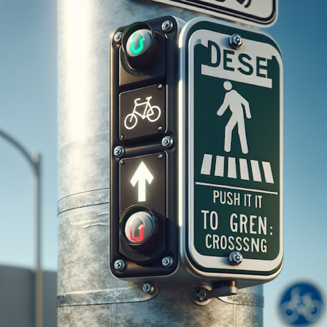} \\
        &
        \tiny street signs at the intersection of cabot road and college avenue under an overcast sky & 
        \tiny stainless steel elevator in an outdoor station with assistance button and sign, ready for use &
        \tiny pedestrian crossing button with directional arrow and qr code for safe street crossing instructions \\
        \raisebox{0\height}{\rotatebox{90}{\shortstack{\small \textbf{out-of-training} \\ {\scriptsize ~}}}} & 
        \includegraphics[width=0.11\textwidth]{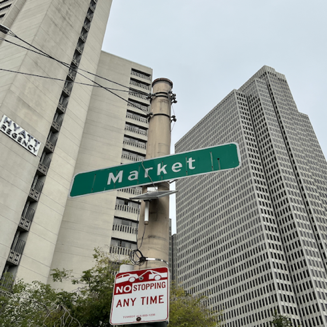} & 
        \includegraphics[width=0.11\textwidth]{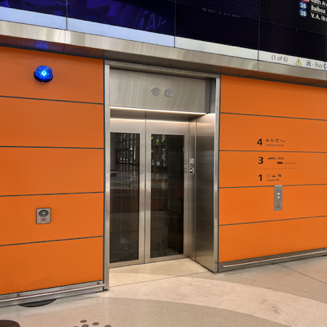} & 
        \includegraphics[width=0.11\textwidth]{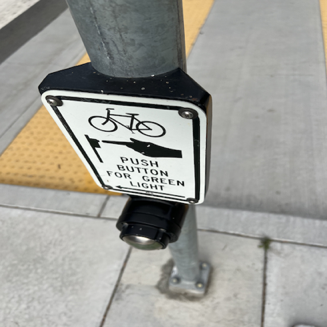} \\    
        &
        \tiny market street sign and no stopping sign against towering concrete and glass buildings on a cloudy day  & 
        \tiny modern stainless steel elevator with orange walls and floor number indicators at a transit station & 
        \tiny bicycle crossing signal button on a pole, instructing cyclists to push for green light \\
        %
        
        %
    \end{tabular}
    }
    \begin{tabular}{c} \\
    \end{tabular}
    \fbox{
        \begin{tabular}{p{0.02\textwidth}p{0.11\textwidth}@{\hspace{8pt}}p{0.11\textwidth}@{\hspace{8pt}}p{0.11\textwidth}}
        \raisebox{0.15\height}{\rotatebox{90}{\shortstack{\small \textbf{in-training} \\ {\scriptsize ~}}}} & 
        \includegraphics[width=0.11\textwidth]{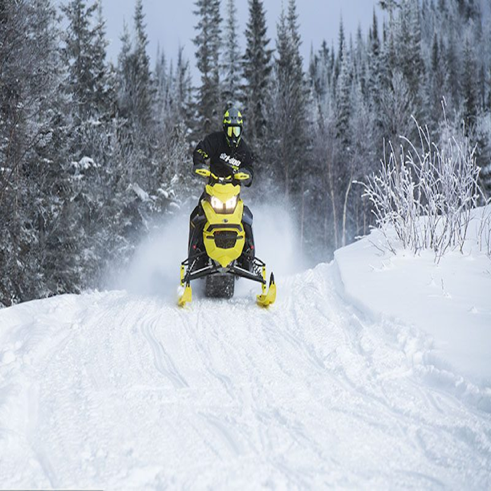} & 
        \includegraphics[width=0.11\textwidth]{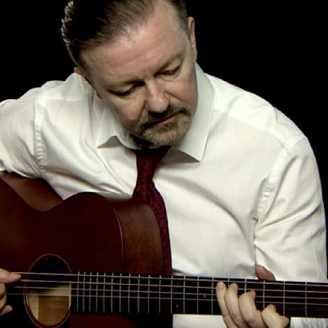} & 
        \includegraphics[width=0.11\textwidth]{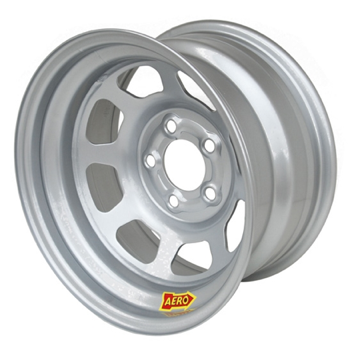} \\
        \raisebox{0\height}{\rotatebox{90}{\shortstack{\small \textbf{out-of-training} \\ {\scriptsize (DALL-E)}}}} & 
        \includegraphics[width=0.11\textwidth]{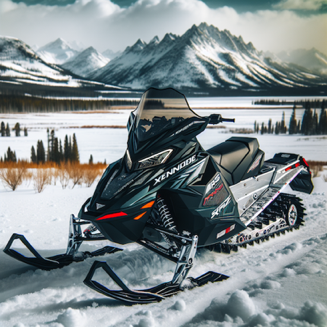} & 
        \includegraphics[width=0.11\textwidth]{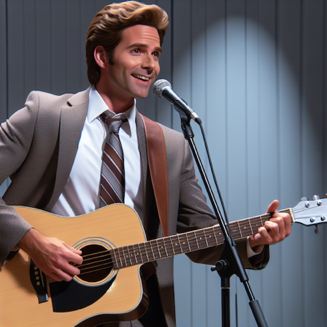} &
        \includegraphics[width=0.11\textwidth]{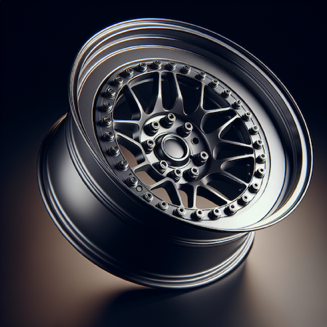} \\ 
        &
        \tiny{2022 Ski-Doo Renegade X-RS 900 ACE Turbo R ES RipSaw 1.25 w/ Premium Color Display in Ponderay, Idaho - Photo 5} &
        \tiny{Listen to Ricky Gervais Perform ``Slough'' as David Brent}  &
        \tiny{Aero 50-074535 50 Series 15x7 Inch Wheel, 5 on 4-1/2 BP 3-1/2 Inch BS} \\
    \end{tabular}
    }
    \begin{tabular}{c} \\
    \end{tabular}
    \fbox{
    \begin{tabular}{p{0.02\textwidth}p{0.11\textwidth}@{\hspace{8pt}}p{0.11\textwidth}@{\hspace{8pt}}p{0.11\textwidth}}
        \raisebox{0.15\height}{\rotatebox{90}{\shortstack{\small \textbf{in-training} \\ {\scriptsize ~}}}} & 
        \includegraphics[width=0.11\textwidth]{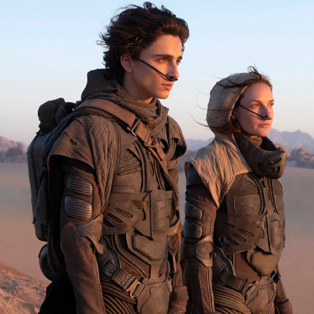} & 
        \includegraphics[width=0.11\textwidth]{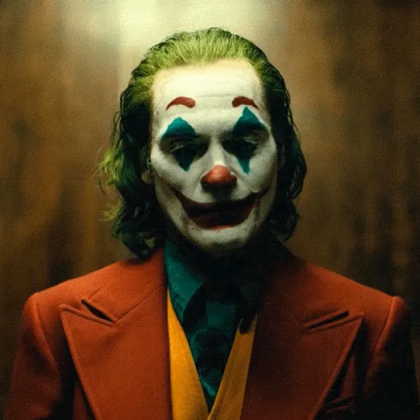} & 
        \includegraphics[width=0.11\textwidth]{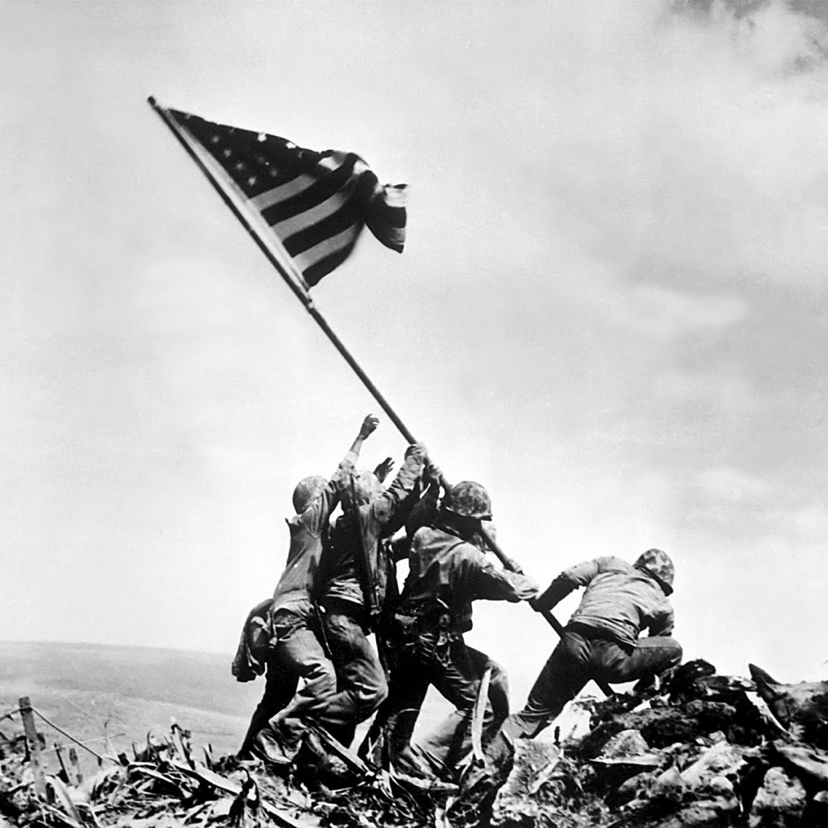} \\
        \raisebox{0\height}{\rotatebox{90}{\shortstack{\small \textbf{out-of-training} \\ {\scriptsize (SD (v3))}}}} & 
        \includegraphics[width=0.11\textwidth]{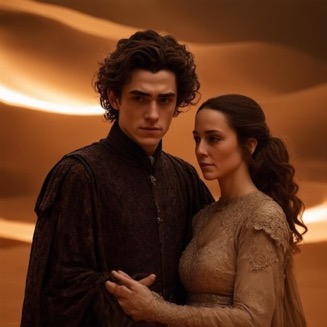} & 
        \includegraphics[width=0.11\textwidth]{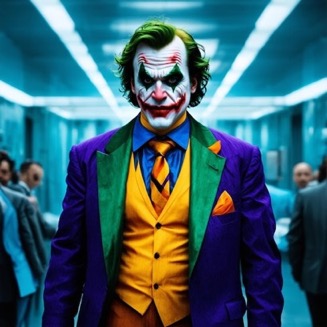} & 
        \includegraphics[width=0.11\textwidth]{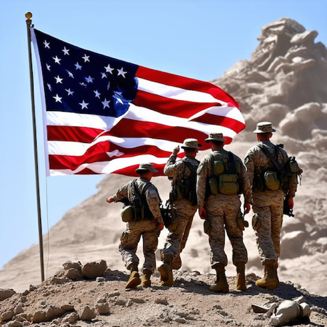} \\
        &
        \tiny{Timothee Chalamet and Rebecca Ferguson in Dune} &
        \tiny{Joker Joaquin Phoenix 2019 Movie Wallpaper Joker} &
        \tiny{Raising the Flag on Iwo Jima} \\
    \end{tabular}
    }
    \vspace{0.2cm}
    \caption{Representative examples of in-training and out-of-training images from the STROLL (top), Carlini (middle), and Midjourney (bottom) datasets.}
    \label{fig:example-images}
\end{figure}

\subsubsection{STROLL} 

This dataset contains $100$ in-training and out-of-training image pairs of outdoor city objects and scenes recorded on a smartphone in the San Francisco Bay area over the course of two days in July 2024. Shown in the top panel of Figure~\ref{fig:example-images} are representative image pairs (first two rows). 

Prompted with ``Provide a detailed, $15$ word long caption of this image,'' ChatGPT-4o~\cite{chatgpt4o} was used to generate a detailed caption for each image. These captions were used for the in-training/out-of-training experiment. For a second, in-training (alt caption)/out-of-training experiment, a new caption was generated for the in-training image (with the out-of-training captions remaining the same). This alternate caption was generated with the prompt ``Provide a detailed, $15$ word long caption of this image that is distinctly different from [previous caption]''.

\subsubsection{Carlini} 

This dataset contains $74$ images that appear to have been memorized~\cite{carlini2023extracting} by Stable Diffusion (v1.4)~\cite{rombach2022high}. Shown in the middle panel of Figure~\ref{fig:example-images} are representative images. These images are a tiny fraction of the LAION-5B~\cite{schuhmann2022laion} dataset used to train Stable Diffusion (v1.4). Each in-training image in this dataset is accompanied by its original caption from LAION-5B, which was also used to generate matching out-of-training images using DALL-E (v2)~\cite{ramesh2022hierarchical} (middle panel of Figure~\ref{fig:example-images}).

\subsubsection{Midjourney} 

This dataset contains $10$ images that appear to have been memorized~\cite{thompson2024jokerai} by Midjourney (v6)~\cite{midjourney}. Shown in the bottom panel of Figure~\ref{fig:example-images} are representative images. Each in-training image in this dataset is accompanied by its original caption from LAION-5B, which was used to generate matching out-of-training images using Stable Diffusion (v3)~\cite{esser2024scaling} (Figure~\ref{fig:example-images}). Unlike the previous two datasets in which the control images were generated using DALL-E, here we use Stable Diffusion (v3) because DALL-E would not generate many of the images in this dataset consisting of recognizable celebrities.

\subsection{Models}

\subsubsection{STROLL/Stable Diffusion (v2.1)}

We created a custom derivative of the Stable Diffusion (v2.1) variant image model by fine-tuning the 2.1 model weights\footnote{\scriptsize{\href{http://huggingface.co/stabilityai/stable-diffusion-2-1}{\texttt{www.huggingface.co/stabilityai/stable-diffusion-2-1}}}} on the in-training portion of the STROLL dataset. The official training script\footnote{\scriptsize\href{https://github.com/huggingface/diffusers/blob/main/examples/text_to_image/train_text_to_image.py}{%
\begingroup\ttfamily\hyphenchar\font=`\-\relax
https://github.com\allowbreak/huggingface\allowbreak/diffusers\allowbreak/blob\allowbreak/main\allowbreak/examples\allowbreak/text\_to\_image\allowbreak/train\_text\_to\_image.py%
\endgroup}} was used to fine-tune the model's UNet module while keeping the CLIP encoder and variational autoencoder (VAE) weights frozen. The learning rate was set to $10^{-5}$, and the maximum-steps parameter was set to $100$\footnote{For reference, fine-tuning SD v2.1 from v2.0 took $210{,}000$ steps.}, while the remaining parameters were left at their recommended default values: image resolution of $512 \times 512$, mixed precision turned off, and a random horizontal flip augmentation. 

Given a seed image and text prompt as input, the image-to-image feature of this fine-tuned model, with default parameters and varying strengths, was used to power our membership inference. The strength $s_i \in [0,1]$ controls the influence of the text prompt relative to the seed image, where a value of $0$ yields a generated image that is identical to the seed image, and a value of $1$ generates an image guided fully by the text prompt, effectively ignoring the seed image.

In order to determine the impact of the number of training steps, we created a second fine-tuned model in which the training steps was increased from $100$ to $1{\tiny,}000$.
    
\subsubsection{Carlini/Stable Diffusion (v1.4)} 

We used the off-the-shelf Stable Diffusion (v1.4) model\footnote{\scriptsize{\href{https://huggingface.co/CompVis/stable-diffusion-v1-4}{\texttt{www.huggingface.co/CompVis/stable-diffusion-v1-4}}}} and invoked its image-to-image pipeline. All image generation parameters were set to the default values with image strength $s_i \in [0,1]$.

\subsubsection{Midjourney/Midjourney (v6)}

We used the commercial Midjourney model\footnote{\scriptsize{\href{https://docs.midjourney.com/docs/image-prompts}{\texttt{docs.midjourney.com/docs/image-prompts}}}} and manually invoked its image-to-image pipeline through their Discord interface. All parameters were set to the default values, except for the image strength (termed weight in Midjourney), ranging from $0$ (yielding a generated image that ignores the seed image) to $3$ (yielding a generated image identical to the seed image). Note that this strength parameter is reversed as compared to Stable Diffusion.

\subsection{Membership Inference}
\label{subsec:membership_inference}

Our membership inference method predicts whether a model $M$ was trained on an image $I$ with caption $C$. This method does not access any explicit information about $M$'s architecture or trained weights. This method only requires access to the image-to-image inference engine for generating an image from a descriptive prompt, seed image, and variable strength parameter that controls the deviation between the seed image and generated image. Intuitively, this approach exploits a -- perhaps unintended -- property of image-to-image generation that produces less variation for an in-training seed image as compared to an out-of-training seed image.

Our method involves three steps: (1) image-to-image inference with varying strengths, (2) measurement of perceptual similarity between a generated and seed image $I$; and (3) membership inference prediction quantifying the likelihood that model $M$ was trained on image $I$.

In the first step, the image-to-image pipeline of model $M$ is invoked with a seed image $I$, its descriptive caption $C$, and strength parameters $s_i$, where $i=1,2,\ldots,m$. For each strength $s_i$, the image generation is repeated $n$ times, resulting in a set of output images $\hat{I}_{i,j}$, where $j=1,2,\ldots,n$. 

In the second step, the distance $d_{i,j}$ between the seed image $I$ and each generated image $\hat{I}_{i,j}$ is calculated using DreamSim~\cite{fu2024dreamsim}. This perceptual metric of image similarity computes a distance $d_{i,j} \in [0,1]$ where a value of $0$ is maximally similar and a value of $1$ is maximally different. For each strength $s_i$, the minimum distance across $n$ generated images is retained, yielding a $m$-D vector of distances for each strength value: $\vec{d} = \begin{pmatrix} d_1 & d_2 & \ldots & d_m \end{pmatrix}$. 

We use a simple logistic-regression model to distinguish in-training from out-of-training images based on the distances $\vec{d}$. 

Stable Diffusion and Midjourney afford a different parametrization of the strength variable $s_i$: For Stable Diffusion (v1.4 and v2.1), $s_i \in [0.02, 0.2, 0.4, 0.6, 0.8, 1.0]$; for Midjourney (v6), $s_i \in [0,1,2,3]$. For Stable Diffusion, we used a minimum strength of $0.02$ because a strength of $0.0$ simply returned the seed image. Throughout, $n=10$ were generated at each strength parameter.

\section{Results}
\label{sec:membership-inference}

As described in detail in Section~\ref{sec:methods}, our membership inference method predicts whether a model $M$ was trained on an image $I$ with caption $C$. This method involves three steps: (1) image-to-image inference with varying strengths, (2) measurement of perceptual similarity between a generated image and seed image $I$ with caption $C$; and (3) membership inference prediction quantifying the likelihood that model $M$ was trained on image $I$. Intuitively, this method exploits an emergent property in which image-to-image generation produces less variation for an in-training seed image as compared to an out-of-training seed image.

\setlength{\fboxrule}{0.5pt}
\setlength{\fboxsep}{3pt} 
\begin{figure}[p]
    \centering
    \begin{tabular}{c}
        \fbox{\includegraphics[width=0.8\linewidth]{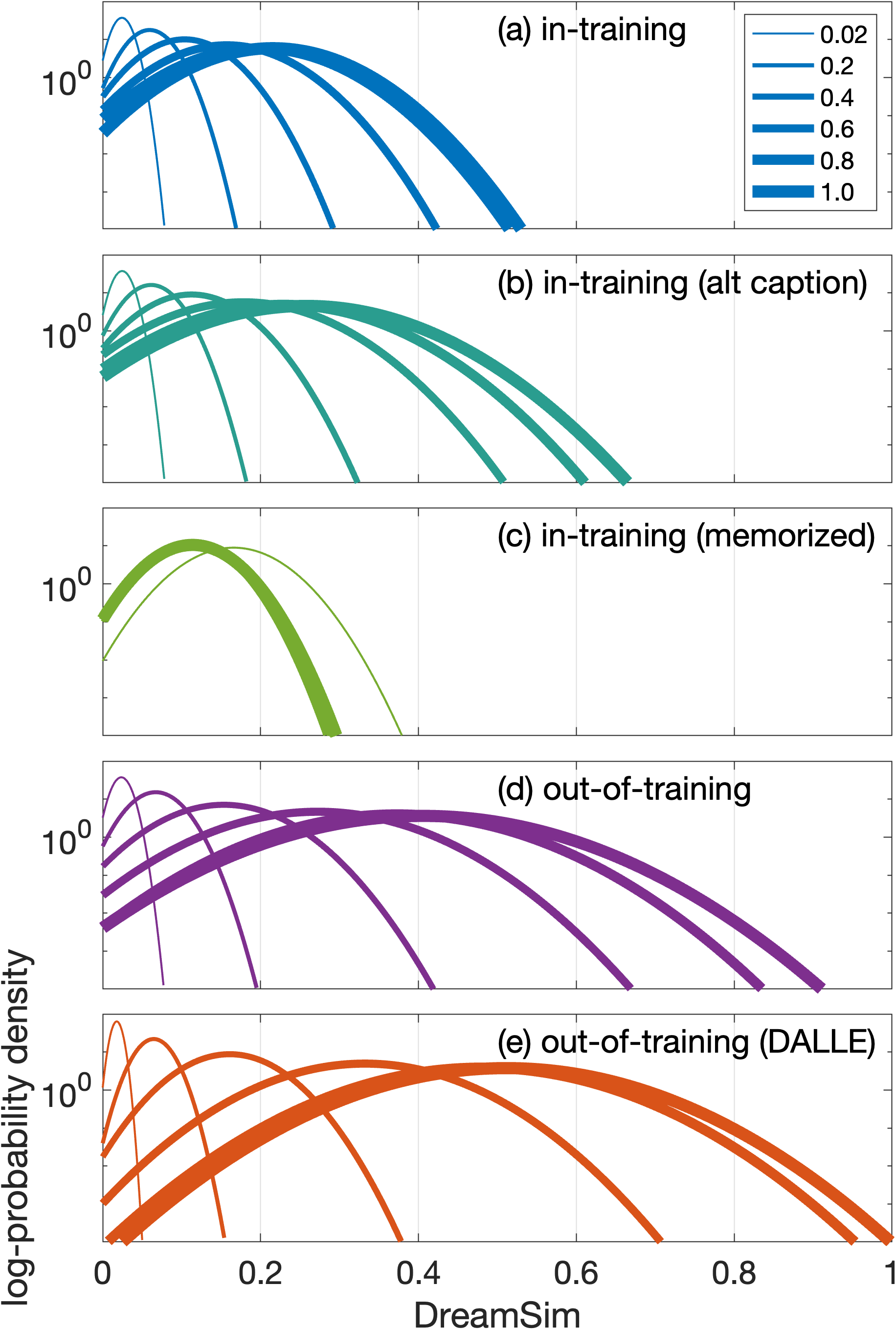}} \\ \\
        \fbox{\includegraphics[width=0.8\linewidth]{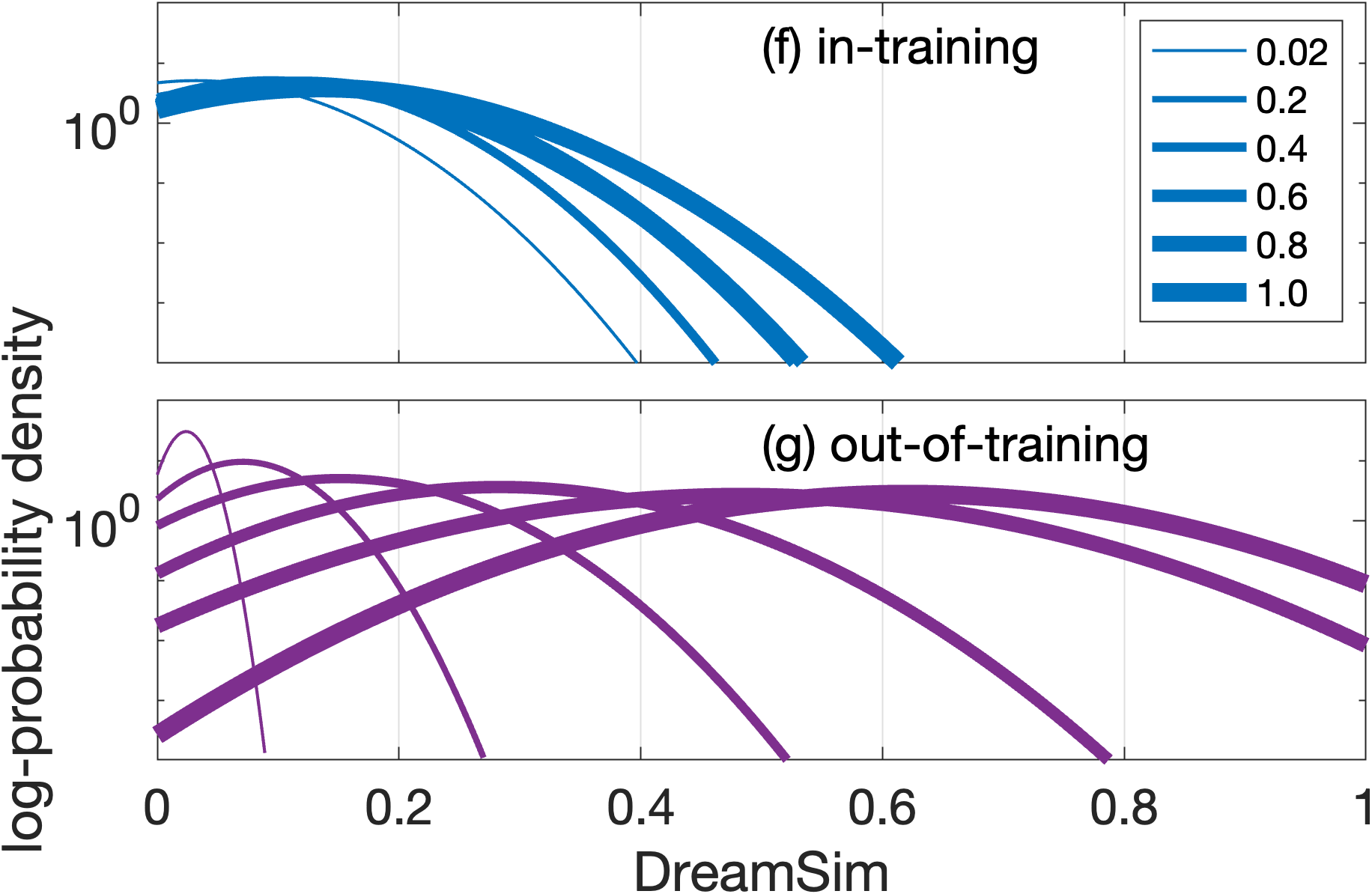}} \\ \\
        \fbox{\includegraphics[width=0.8\linewidth]{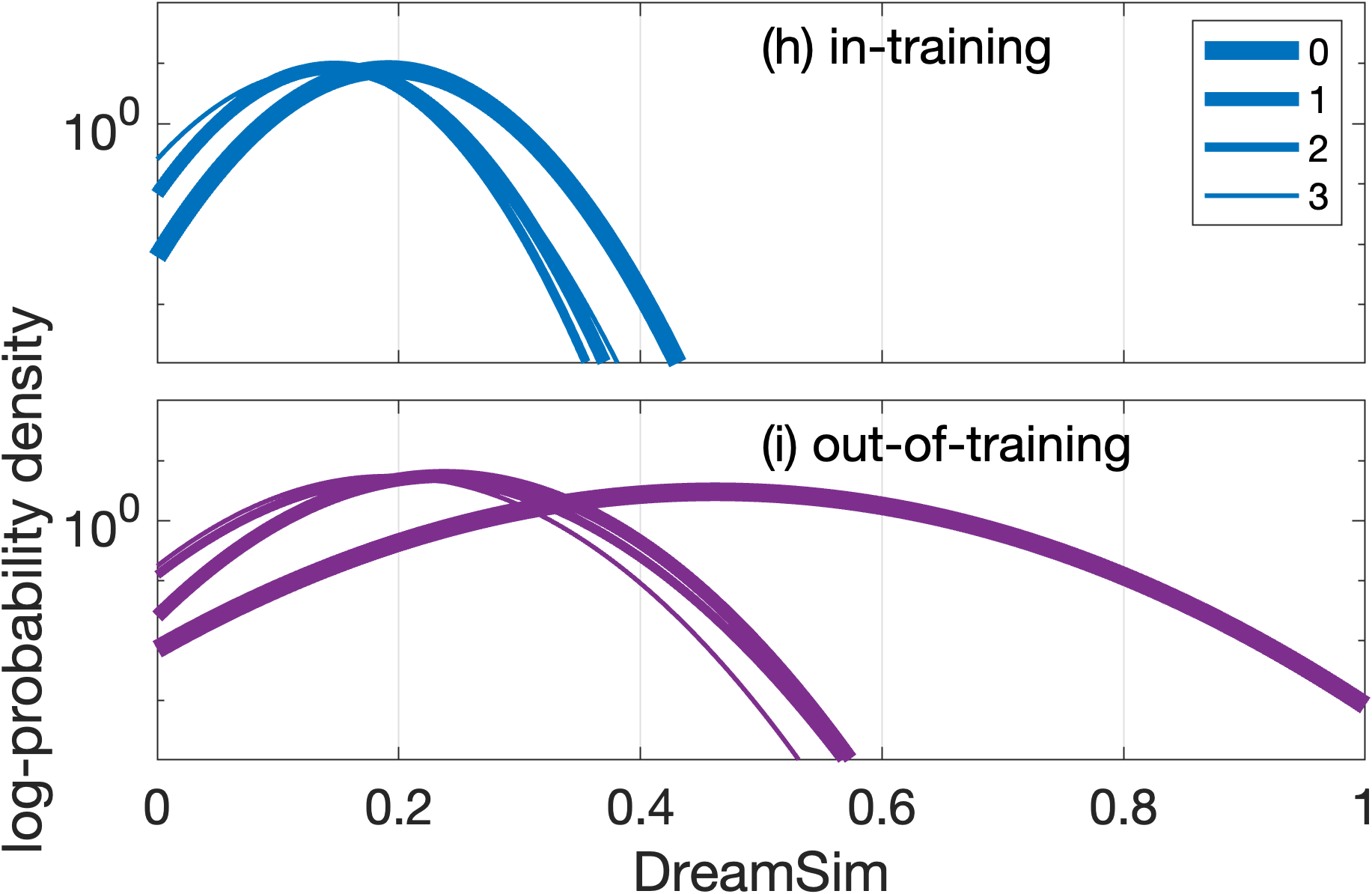}}
        \end{tabular}
    \caption{The log probability densities for image similarity (DreamSim) between a seed image and a generated image with varying strength for the STROLL  (top), Carlini (middle), and Midjourney (bottom) datasets. In all cases, in-training seed images lead to generated images that are more perceptually similar to the seed (smaller DreamSim distance).}
    \label{fig:results-logprob}
\end{figure}

\subsection{STROLL}

Shown in Figure~\ref{fig:results-logprob}(a) is a Gaussian fitted log-probability density function to the DreamSim distance between the in-training seed images and the result of image-to-image generation under Stable Diffusion 2.1. Each curve corresponds to a different image-to-image strength parameter $s_i$ (see Section~\ref{sec:methods}) where, as strength increases, the seed image has increasingly less impact on the generated image. As expected, for strength parameters close to $0$ (thinnest curve), the DreamSim distance is relatively small, and as the strength increases (thicker curves), the distance increases proportionally, meaning that the generated images are increasingly more distinct from the seed image.

Shown in Figure~\ref{fig:results-logprob}(d) are the same density functions but for the out-of-training seed images. Here we see the same trend, where small strength parameters lead to more similarity as compared to larger strength parameters. However, the mean of these densities as a function of strength $s_i$ is larger for these out-of-training images. In particular, notice that the mean of the densities for strengths greater than $0.6$ are significantly larger for out-of-training as compared to in-training images. That is, the images generated with an out-of-training seed are more distinct than those generated with an in-training seed.

\begin{figure*}[p!]
    \centering

    \resizebox{!}{0.29\textwidth}{%
    \fbox{
        \begin{tabular}{p{0.01\textwidth}p{0.12\textwidth}@{\hspace{10pt}}p{0.12\textwidth}@{\hspace{4pt}}p{0.12\textwidth}@{\hspace{4pt}}p{0.12\textwidth}@{\hspace{4pt}}p{0.12\textwidth}@{\hspace{4pt}}p{0.12\textwidth}@{\hspace{4pt}}p{0.12\textwidth}@{\hspace{4pt}}}
            & \centering seed & \centering 0.02 & \centering 0.2 & \centering 0.4 & \centering 0.6 & \centering 0.8 & \multicolumn{1}{c}{1.0} \\
    
            \raisebox{0.24\height}{\rotatebox{90}{\shortstack{\small \textbf{in-training}}}} & 
            \includegraphics[width=0.12\textwidth]{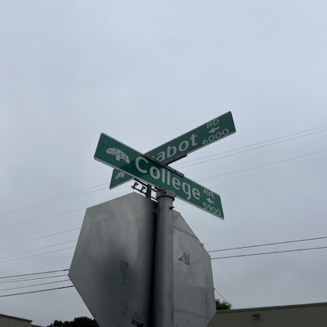} & 
            \includegraphics[width=0.12\textwidth]{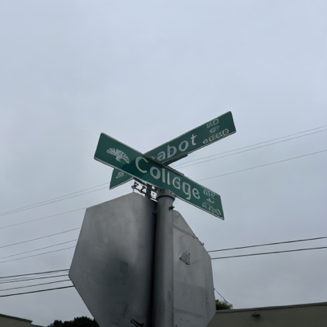} & 
            \includegraphics[width=0.12\textwidth]{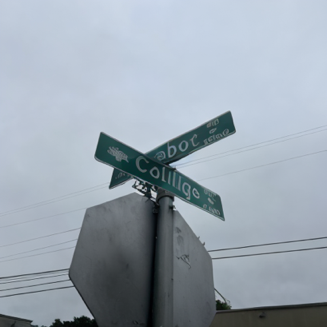} & 
            \includegraphics[width=0.12\textwidth]{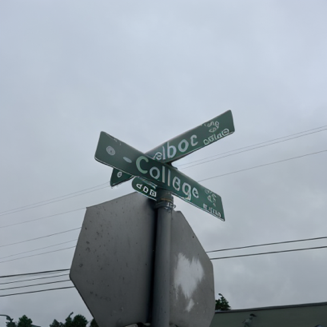} & 
            \includegraphics[width=0.12\textwidth]{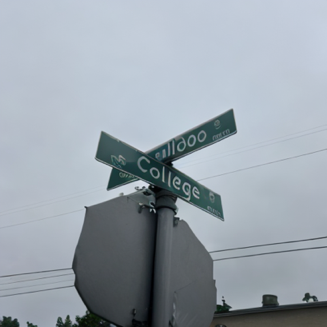} & 
            \includegraphics[width=0.12\textwidth]{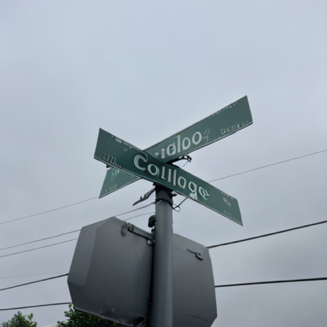} & 
            \includegraphics[width=0.12\textwidth]{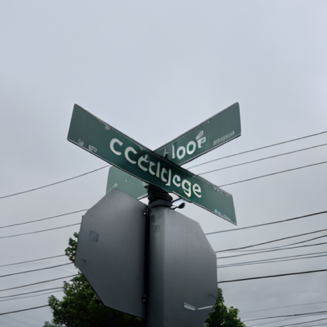} \\
    
            \raisebox{0.03\height}{\rotatebox{90}{\shortstack{\small \textbf{out-of-training}}}} & 
            \includegraphics[width=0.12\textwidth]{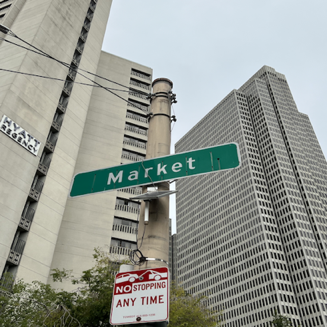} & 
            \includegraphics[width=0.12\textwidth]{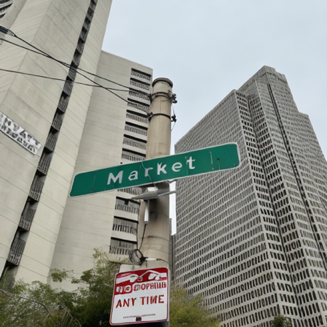} & 
            \includegraphics[width=0.12\textwidth]{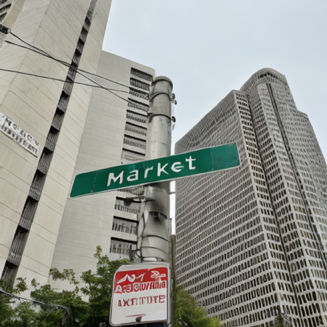} & 
            \includegraphics[width=0.12\textwidth]{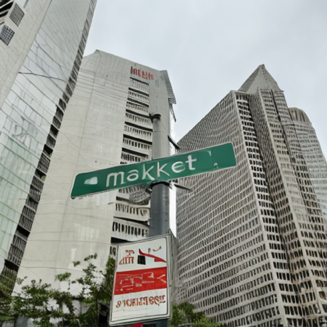} & 
            \includegraphics[width=0.12\textwidth]{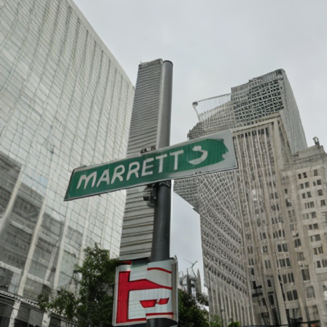} & 
            \includegraphics[width=0.12\textwidth]{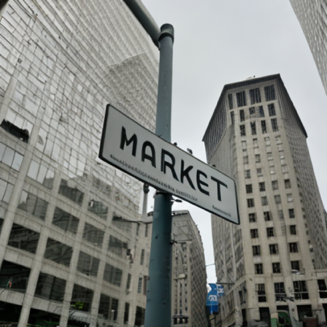} & 
            \includegraphics[width=0.12\textwidth]{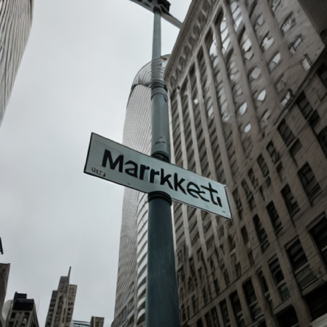} \\ \\
    
            \raisebox{0.24\height}{\rotatebox{90}{\shortstack{\small \textbf{in-training}}}} & 
            \includegraphics[width=0.12\textwidth]{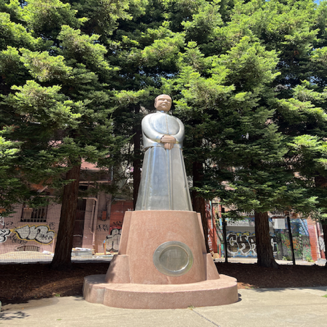} & 
            \includegraphics[width=0.12\textwidth]{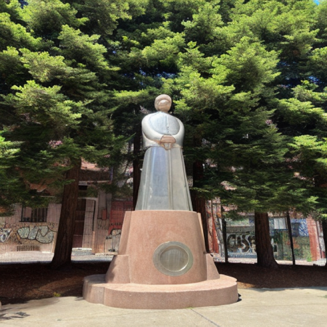} & 
            \includegraphics[width=0.12\textwidth]{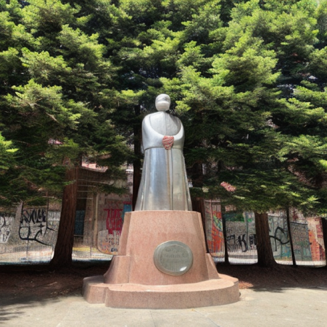} & 
            \includegraphics[width=0.12\textwidth]{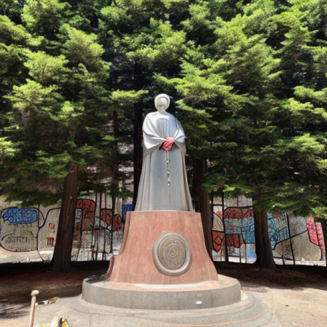} & 
            \includegraphics[width=0.12\textwidth]{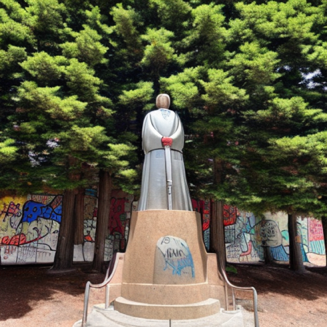} & 
            \includegraphics[width=0.12\textwidth]{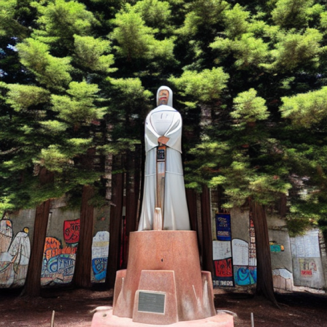} & 
            \includegraphics[width=0.12\textwidth]{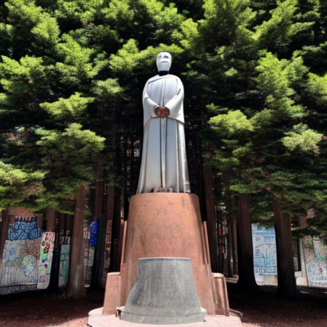} \\
    
            \raisebox{0.03\height}{\rotatebox{90}{\shortstack{\small \textbf{out-of-training}}}} & 
            \includegraphics[width=0.12\textwidth]{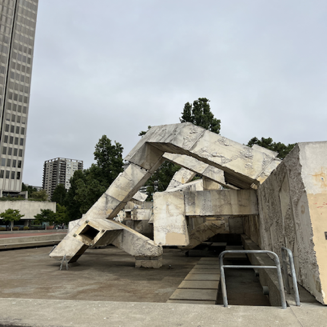} & 
            \includegraphics[width=0.12\textwidth]{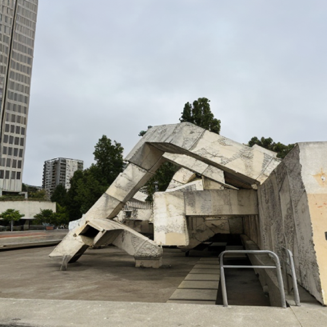} & 
            \includegraphics[width=0.12\textwidth]{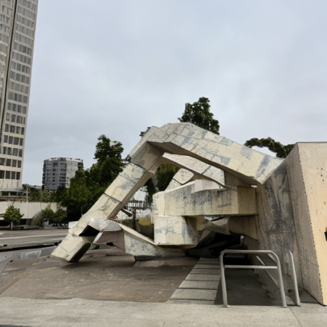} & 
            \includegraphics[width=0.12\textwidth]{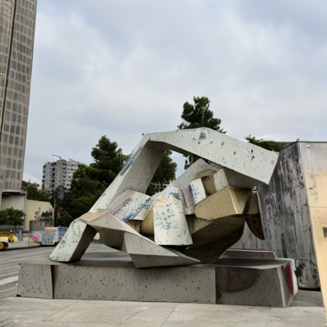} & 
            \includegraphics[width=0.12\textwidth]{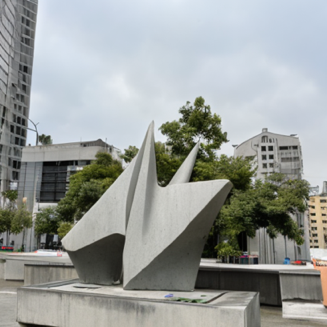} & 
            \includegraphics[width=0.12\textwidth]{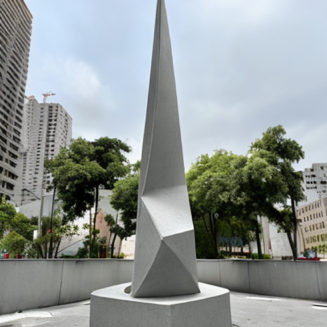} & 
            \includegraphics[width=0.12\textwidth]{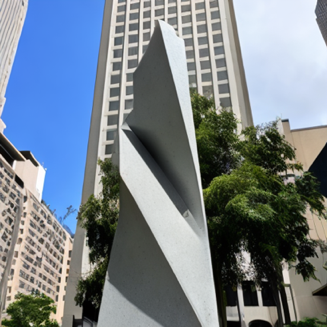}
        \end{tabular}
        }}

    \begin{tabular}{c}
    ~~~~~\\
    \end{tabular}

    \resizebox{!}{0.15\textwidth}{%
    \fbox{
        \begin{tabular}{p{0.01\textwidth}p{0.12\textwidth}@{\hspace{10pt}}p{0.12\textwidth}@{\hspace{4pt}}p{0.12\textwidth}@{\hspace{4pt}}p{0.12\textwidth}@{\hspace{4pt}}p{0.12\textwidth}@{\hspace{4pt}}p{0.12\textwidth}@{\hspace{4pt}}p{0.12\textwidth}@{\hspace{4pt}}}
            & \centering seed & \centering 0.02 & \centering 0.2 & \centering 0.4 & \centering 0.6 & \centering 0.8 & \multicolumn{1}{c}{1.0} \\
    
            \raisebox{0.24\height}{\rotatebox{90}{\shortstack{\small \textbf{in-training}}}} & 
            \includegraphics[width=0.12\textwidth]{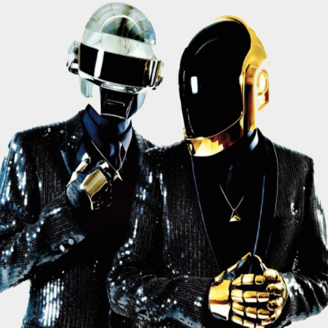} & 
            \includegraphics[width=0.12\textwidth]{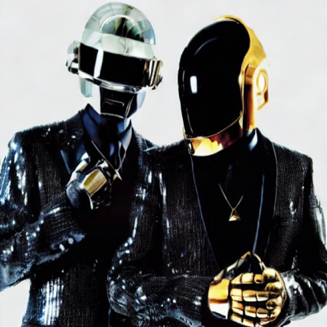} & 
            \includegraphics[width=0.12\textwidth]{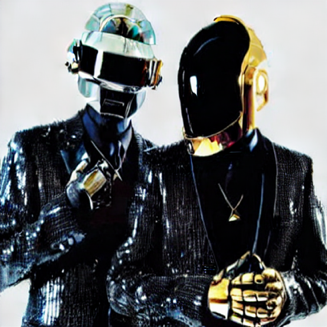} & 
            \includegraphics[width=0.12\textwidth]{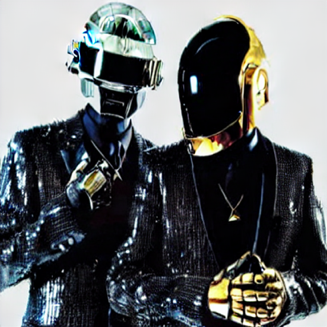} & 
            \includegraphics[width=0.12\textwidth]{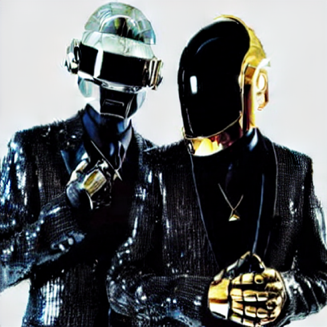} & 
            \includegraphics[width=0.12\textwidth]{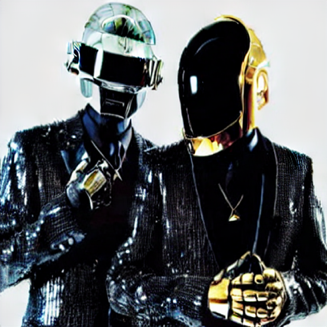} & 
            \includegraphics[width=0.12\textwidth]{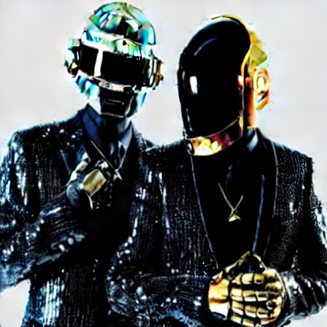} \\
    
            \raisebox{0.03\height}{\rotatebox{90}{\shortstack{\small \textbf{out-of-training}}}} & 
            \includegraphics[width=0.12\textwidth]{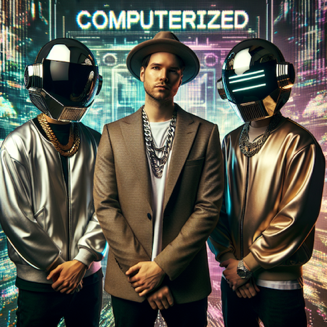} & 
            \includegraphics[width=0.12\textwidth]{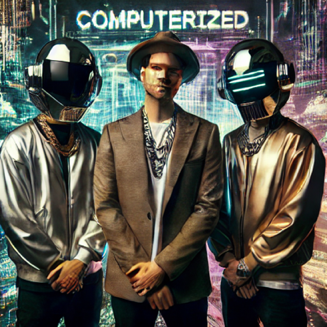} & 
            \includegraphics[width=0.12\textwidth]{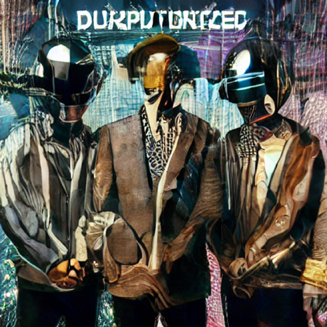} & 
            \includegraphics[width=0.12\textwidth]{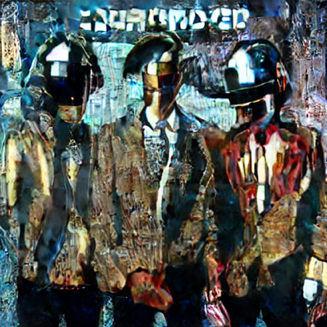} & 
            \includegraphics[width=0.12\textwidth]{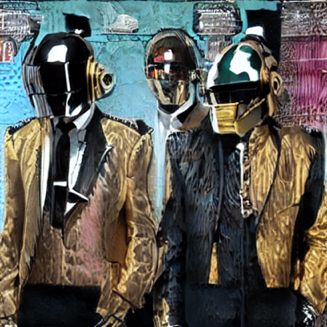} & 
            \includegraphics[width=0.12\textwidth]{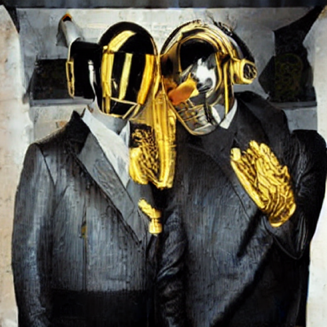} & 
            \includegraphics[width=0.12\textwidth]{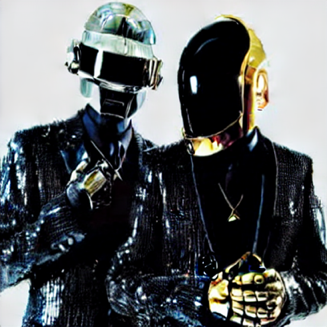}
        \end{tabular}
    }}

    \begin{tabular}{c}
    ~~~~~\\
    \end{tabular}
    
    \resizebox{!}{0.15\textwidth}{%
    \fbox{
        \begin{tabular}{p{0.01\textwidth}p{0.12\textwidth}@{\hspace{10pt}}p{0.12\textwidth}@{\hspace{4pt}}p{0.12\textwidth}@{\hspace{4pt}}p{0.12\textwidth}@{\hspace{4pt}}p{0.12\textwidth}}
            & \centering seed & \centering 3 & \centering 2 & \centering 1 & \multicolumn{1}{c}{0} \\
    
            \raisebox{0.24\height}{\rotatebox{90}{\shortstack{\small \textbf{in-training}}}} & 
            \includegraphics[width=0.12\textwidth]{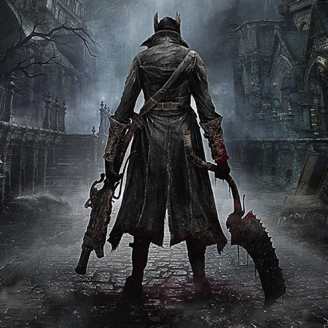} & 
            \includegraphics[width=0.12\textwidth]{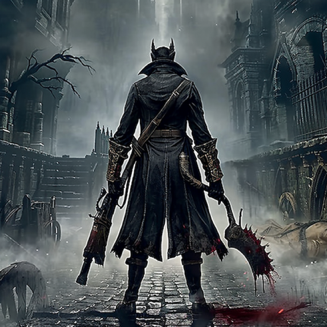} & 
            \includegraphics[width=0.12\textwidth]{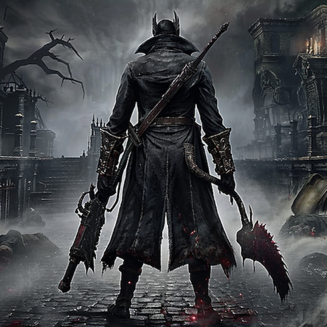} & 
            \includegraphics[width=0.12\textwidth]{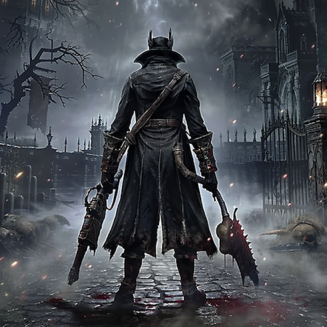} & 
            \includegraphics[width=0.12\textwidth]{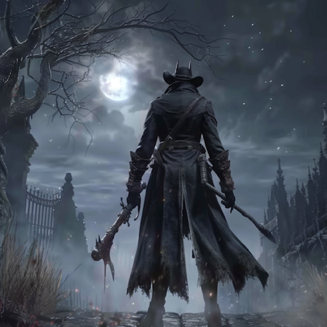} \\
    
            \raisebox{0.03 \height}{\rotatebox{90}{\shortstack{\small \textbf{out-of-training}}}} & 
            \includegraphics[width=0.12\textwidth]{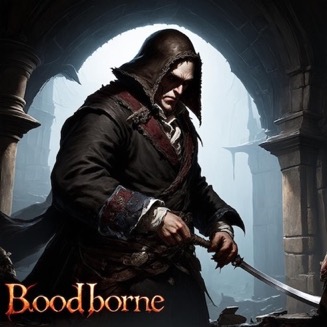} & 
            \includegraphics[width=0.12\textwidth]{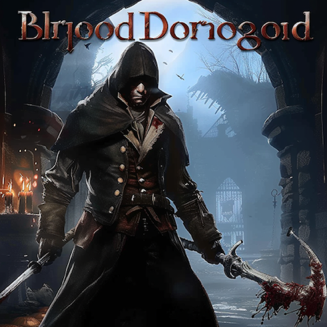} & 
            \includegraphics[width=0.12\textwidth]{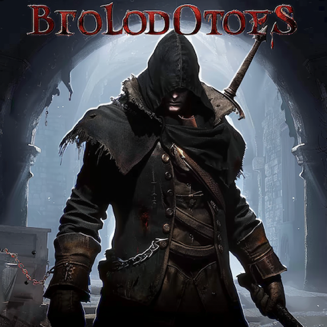} & 
            \includegraphics[width=0.12\textwidth]{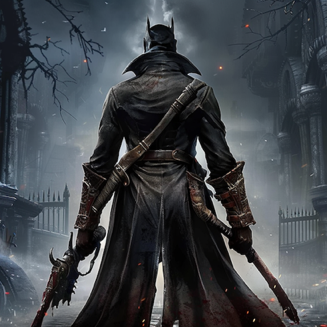} & 
            \includegraphics[width=0.12\textwidth]{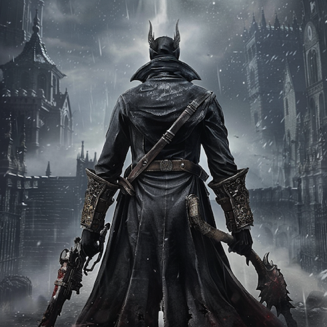}
        \end{tabular}
        \hspace{4.36cm}
    }}
    \caption{Images generated using the STROLL dataset and Stable Diffusion (v2.1) image-to-image pipeline (top), the Carlini dataset and Stable Diffusion (v1.4) model (middle), and Midjourney (v6) dataset and image-to-image pipeline. The seed image is shown in the left-most column. The remaining columns correspond to generated images with increasing strength, where moving rightward corresponds to less emphasis placed on the seed image. In all three cases, the in-training seed image leads to perceptually more similar images than the out-of-training seed images.}
    \label{fig:results-images}
\end{figure*}

Shown in the top panel of Figure~\ref{fig:results-images} is an in-training seed image (far left) and the resulting image-to-image generation for strengths $s_i \in [0.02, 0.2, 0.4, 0.6, 0.8, 1.0]$. Consistent with the DreamSim distance (Figure~\ref{fig:results-logprob}(a)), all of the generated images are perceptually similar to the seed image. By comparison, also shown in Figure~\ref{fig:results-images}, is an out-of-training seed image and the resulting image-to-image generation for varying strengths. Again, consistent with the DreamSim distance (Figure~\ref{fig:results-logprob}(d)), the generated images deviate from the seed starting at a strength of $0.6$.

Independent-samples, two-sided t-tests reveal a significant difference between the DreamSim distributions for the in-training vs out-of-training data at a strength $s_i \ge 0.4$, with an increasing effect size (Cohen's D) with increasing strength:
\begin{itemize} [itemsep=-0.1em]
    \item $0.02$ ($t(198)=0.5,~p=0.6,~D=0.07$)
    \item $0.2$ ($t(198)=2.3,~p=0.02,~D=0.32$)
    \item $0.4$ ($t(198)=7.3,~p<10^{-8},~D=1.0$)
    \item $0.6$ ($t(198)=11.2,~p<10^{-8},~D=1.6$)
    \item $0.8$ ($t(198)=14.5,~p<10^{-8},~D=2.1$)
    \item $1.0$ ($t(198)=14.8,~p<10^{-8},~D=2.1$)
\end{itemize}

\subsubsection{Alternate Caption}

In the above analysis, we assume that the image-to-image generation is provided with the same image caption used in training of the image-generation model. To test the sensitivity of this assumption, alternate captions were generated for each image (see Section~\ref{sec:methods}). Shown in  Figure~\ref{fig:results-logprob}(b) are the same Gaussian fitted log-probability density functions to the DreamSim distances. As compared to the in-training images with the original caption (Figure~\ref{fig:results-logprob}(a)), the generated images with alternate captions are less similar to their seed images, but still distinct from the out-of-training images (Figure~\ref{fig:results-logprob}(d)).

An independent-samples, two-sided t-tests again reveals a significant difference between the DreamSim distributions for the in-training vs out-of-training data at a strength $s_i \ge 0.4$:
\begin{itemize} [itemsep=-0.1em]
    \item $0.02$ ($t(198)=0.7,~p=0.5, D=0.1$)
    \item $0.2$ ($t(198)=1.6,~p=0.1, D=0.2$)
    \item $0.4$ ($t(198)=5.7,~p<10^{-8},~D=0.8$)
    \item $0.6$ ($t(198)=8.5,~p<10^{-8},~D=1.2$) 
    \item $0.8$ ($t(198)=11.1,~p<10^{-8},~D=1.6$)
    \item $1.0$ ($t(198)=10.5,~p<10^{-8},~D=1.5$)
\end{itemize}

\subsubsection{Effect Size}

We observe qualitative differences between the distributions for the in-training and out-of-training data increases with strength $s_i$. We quantify this with Cohen's D, a measure of effect size: for strengths $0.02, 0.2, \ldots, 1.0$, the effect sizes are $0.1, 0.3, 1.0, 1.6, 2.1, 2.1$. A similar pattern emerges for the in-training and out-of-training (alt caption) distributions, with effect sizes of $0.1, 0.2, 0.8, 1.2, 1.6, 1.5$. With a Cohen's D of $0.8$ considered large, we see large effects with strength parameters greater than $0.4$

\subsubsection{Classifier}

The distributions in the top panel Figure~\ref{fig:results-logprob} show a population-level difference between in-training and out-of-training images. To predict if an individual image belongs to the in-training or out-of-training set, we trained a logistic regression on the $6$-D distance vectors $\vec{d}$ corresponding to the $100$ in-training (original caption), $100$ in-training (alternate caption), $100$ out-of-training and $100$ out-of-training (DALL-E) images. These DALL-E images, generated to match the content of the out-of-training images (see Section~\ref{sec:methods}), balanced the data set for model training. As shown in Figure~\ref{fig:results-logprob}(e), the distributions for these images are similar to the out-of-training images in panel (d).

A logistic regression was trained on a random subset of $80\%$ of this data and evaluated on the remaining $20\%$. From $100$ random training/testing splits, the average testing accuracy (measured as equal error rate) is $85\%$ with a variance of $0.17$. For a fixed false positive rate (misclassifying an out-of-training image as in-training) of $1\%$, the average true positive rate (correctly classifying an in-training image) is $74\%$ with a variance of $0.36$.

\subsubsection{Memorization}

In the results described above, the fine-tuned Stable Diffusion model was trained with $100$ steps. We next consider the impact of increasing the training steps to $1000$ as described in Section~\ref{sec:methods}. 

Shown in Figure~\ref{fig:results-logprob}(c) is the Gaussian-fitted log probability densities for this model in which we can see a qualitatively different pattern than before, Figure~\ref{fig:results-logprob}(a). Here, almost regardless of strength, the generated images are uniformly similar to the seed image. This, we posit, is because the prolonged learning caused the model to effectively memorize the training images and associated captions. This is consistent with the results described next.

\subsection{Carlini}

In the previous section, we showed that when the pre-trained Stable Diffusion (v2.1) model is fine-tuned on a set of $100$ images of our creation, we can determine that the model was trained on these images. Because this is a fairly constrained experiment, we next validate that our membership inference generalizes to a real-world scenario.

As described in Section~\ref{sec:methods}, the Carlini dataset consists of $74$ images used to train Stable Diffusion (v1.4) and, as shown in~\cite{carlini2023extracting}, this model can be coaxed to produce images that are nearly indistinguishable from these training images, Figure~\ref{fig:example-images}.

Shown in Figure~\ref{fig:results-logprob}(f)-(g) are the DreamSim distances for the $74$ in-training images and $74$ semantically matched out-of-training images (see Section~\ref{sec:methods}). Here, we see the same pattern as with the STROLL results described above: the generated images are perceptually more similar to the in-training seed images than the out-of-training seed images.

Shown in the middle panel of Figure~\ref{fig:results-images} is an in-training seed image (far left) and the resulting image-to-image generation for varying strengths. As with the previous STROLL results, the generated images are perceptually similar to the seed image. By comparison, the out-of-training seed image yields generated images that deviate from the seed starting at a strength of $0.6$. Interestingly, the out-of-training image at strength $s_i = 1.0$ is nearly identical to the in-training seed image. This is because, as shown by Carlini et al.~\cite{carlini2023extracting}, this image was memorized during the original training of the model, and so the original prompt yields the training image at a strength of $1.0$ where the seed image is ignored.

\subsubsection{Classifier}

The average equal error rate for logistic regression trained on the STROLL images and evaluated on the Carlini dataset is $93\%$ with a variance of $0.01$, and for a false positive rate of $1\%$, the average true positive rate is $90\%$ with a variance of $0.01$. This accuracy is somewhat better than for the STROLL images because this dataset was not just trained on but was effectively memorized by the image generator, leading to a larger difference between in-training and out-of-training seed images. Here, we see that the classifier trained on a different dataset and version of Stable Diffusion generalizes quite nicely.

\subsubsection{Memorization}

Note that the in-training distributions, Figure~\ref{fig:results-logprob}(f), are qualitatively similar to the distributions in Figure~\ref{fig:results-logprob}(c) corresponding to the in-training (memorized) results. This, we believe, is because both of these models have memorized some training images and so we see less variation than in the case when the model was simply exposed to these images.

\subsection{Midjourney}
\label{subsec:results_midjourney}

In the previous two sections, we showed the efficacy of our membership inference on two different versions of Stable Diffusion (v2.1 and v1.4). Here we show that our approach generalizes to different model architectures. 

As described in Section~\ref{sec:methods}, the Midjourney dataset consists of $10$ images that appear to have been part of the training datatset for Midjourney (v6). In particular, as shown in~\cite{thompson2024jokerai}, Midjourney can be coaxed to produce images that are nearly indistinguishable from these well-recognized images, Figure~\ref{fig:example-images}.

Shown in Figure~\ref{fig:results-logprob}(h)-(i) are the DreamSim distances for the $10$ in-training images and $10$ semantically matched out-of-training images (see Section~\ref{sec:methods}). Here, we see the same pattern as with the STROLL and Carlini results: the generated images are perceptually more similar to the in-training seed images than the out-of-training seed images.

Shown in the bottom panel of Figure~\ref{fig:results-images} is an in-training seed image (far left) and the resulting image-to-image generation for varying strengths. As with the previous STROLL and Carlini results, the generated images are perceptually similar to the seed image. By comparison, the out-of-training seed image leads to images that deviate more noticeably.

We again see that the out-of-training image, at a strength $s_i = 0$ where the seed image is ignored, is nearly identical to the in-training seed image. This is because this image was memorized during the original training of the model~\cite{thompson2024jokerai}, and so the prompt simply reproduces it.

\subsubsection{Memorization}

As before, the in-training distributions, Figure~\ref{fig:results-logprob}(h), are qualitatively similar to those in Figure~\ref{fig:results-logprob}(c) and (f) corresponding to the STROLL in-training (memorized) and Carlini in-training results. This, again, is because all three of these models have memorized some training images.

\subsubsection{Classifier}

Because Midjourney uses a different strength parametrization than Stable Diffusion, we are not able to deploy the logistic regression model on a per-image basis.

\subsection{Comparison to Previous Work}

Comparing membership inference methods for generative-AI image models remains challenging due to the lack of standardized benchmarks and varying problem definitions/configurations (including the frequency and intensity with which images are presented to the model during training). Nonetheless, we compare our method to existing membership inference approaches for the latest generative-AI image model architectures (diffusion) in terms of computational demands and overall effectiveness.

The method in \cite{wu2022membership} assumes access to a confirmed subset of the targeted model's in-training and out-of-training data. Its reported accuracy ranges from $65\%$ to $100\%$ across multiple models and datasets. The method in \cite{zhai2024membership} assumes access to internal representations of the model during diffusion steps. At a fixed false positive rate of $1\%$, it achieves a true positive rate ranging from $54\%$ to $68\%$ across multiple datasets. Similarly, the method in \cite{li2024unveiling} also relies on internal representations during diffusion steps. At a fixed false positive rate of $1\%$, it achieves a true positive rate between $50\%$ and $58\%$ across different datasets. And, the method in \cite{dubinski2024cdi} is an ensemble of four other methods, whose predictions are analyzed for statistical significance and overall confidence. At a fixed false positive rate of $1\%$, it achieves a true positive rate ranging from $25\%$ to $100\%$ across multiple datasets.

By contrast, our method does not require access to a confirmed subset of in-training/out-of-training data or internal representations from diffusion steps and achieves an average accuracy between $85\%$ and $93\%$. At a fixed false positive rate of $1\%$, our method attains an average true positive rate of between $74\%$ and $90\%$.

When tested in the wild, many existing methods have been found largely ineffective~\cite{duan2024membership}. We, on the other hand, demonstrate that the DreamSim trends leveraged by our method persist even in an in-the-wild setting with Midjourney (v6) (see Section~\ref{subsec:results_midjourney}).

Most effective membership inference and training data extraction methods are computationally expensive. The method in~\cite{carlini2023extracting}, which assumes a white-box scenario with additional access to the model's internals, achieves a true positive rate of $71\%$ at a false positive rate of $1\%$. This approach, however, is computationally demanding, requiring training $16$ shadow models, each of which computes loss for all known in-training data points at each of $1{\small,}000$ diffusion steps. By contrast, our method does not require training and only needs to perform six generations with at most $50$ diffusion steps.

\section{Discussion}

We have observed that when seeded with a previously trained image, image-to-image generation produces an image more similar to the seed image as compared to those generated from an out-of-training seed image. This is distinct from pure memorization, where it has previously been shown that, with a sufficient amount of exposure, models can reproduce training images~\cite{carlini2023extracting}. Our approach applies to both this less common case of memorization as well as the more typical and broader class of training images.

We hypothesize a few different mechanisms that may explain why generative-AI models behave this way. One possibility is that when a seed image is partially corrupted with additive noise (proportional to the user-supplied strength parameter) and placed in the latent space for denoising, because of previous exposure to a training image-caption pair, the previously learned local gradients guide the denoising to a latent representation near a trained image. This explanation would be consistent with the differences seen in Figure~\ref{fig:results-logprob}(a)-(d), where a memorized image/caption yields more self-similar images than an in-training image/caption, which yields more self-similar images than an in-training image/alternate caption. That is, the level of exposure to a specific image/caption pair at training leads to proportionally learned gradients in the denoiser.

Another possibility is that after training, the latent space is non-uniformly structured, and so once an image/caption pair is placed into latent space near an in-training exemplar, it is simply more likely to converge to the in-training image because of this structure. This is more likely to occur with image-to-image generation because the initialization in the latent space is dependent on the seed image and the strength parameter constrains the number of steps that can be taken by the denoiser.

Understanding why models are biased to produce content similar to their training data may provide insights into reducing the likelihood of infringement in the form of reproducing training data, and may provide insights into how a model can be made to forget training exemplars.

An attractive aspect of our membership inference for generative image models is that it does not require access to model architecture details or trained weights, is computationally efficient, and generalizes to multiple different AI models. A drawback of our approach is that it only applies to models that allow for an image-to-image synthesis with a controllable strength parameter, as compared to text-to-image. Depending on the underlying mechanism by which models produce images similar to their in-training data, our method may be adaptable to text-to-image generation.

Many of today's tech leaders have admitted that their generative-AI models would not exist without their training on billions of pieces of content scraped from all corners of the internet~\cite{guardian2024impossible}. These same leaders have also called for the loosening of fair-use and copyright rules. While it is for the courts to decide on these matters of law~\cite{samuelson2023ongoing}, we contend that content creators have legitimate concerns for whether and how their content is used to train generative-AI models, some of which are designed to offer services directly competing with these very content creators.

A critical component of adjudicating these issues will be determining if a deployed model was trained on a specific piece of content. Equally important is determining how creators can and should be compensated when their content is used for training, and how models can be made to forget its training on a specific piece of content should this be the wish of the content's creator. We have focused only on the first of these questions, but all of these issues are important to resolve as generative AI continues its impressive and impactful trajectory.

\section*{Acknowledgments}

This work was funded by funding from the University of California Noyce Initiative. The authors thank Nicholas Carlini for providing the list of extracted instances of memorization from \cite{carlini2023extracting}, and both Nicholas Carlini and Milad Nasr for helpful discussions.

{
    \small
    \bibliographystyle{ieeenat_fullname}
    \bibliography{main}

\begin{thebibliography}{28}
\providecommand{\natexlab}[1]{#1}
\providecommand{\url}[1]{\texttt{#1}}
\expandafter\ifx\csname urlstyle\endcsname\relax
  \providecommand{\doi}[1]{doi: #1}\else
  \providecommand{\doi}{doi: \begingroup \urlstyle{rm}\Url}\fi

\bibitem[cha(2024)]{chatgpt4o}
Chatgpt-4o.
\newblock \url{https://chatgpt.com/}, 2024.

\bibitem[mid(2024)]{midjourney}
Midjourney.
\newblock \url{https://www.midjourney.com/}, 2024.

\bibitem[Carlini et~al.(2023)Carlini, Hayes, Nasr, Jagielski, Sehwag, Tramer, Balle, Ippolito, and Wallace]{carlini2023extracting}
Nicolas Carlini, Jamie Hayes, Milad Nasr, Matthew Jagielski, Vikash Sehwag, Florian Tramer, Borja Balle, Daphne Ippolito, and Eric Wallace.
\newblock Extracting training data from diffusion models.
\newblock In \emph{32nd USENIX Security Symposium (USENIX Security 23)}, pages 5253--5270, 2023.

\bibitem[Duan et~al.(2024)Duan, Suri, Mireshghallah, Min, Shi, Zettlemoyer, Tsvetkov, Choi, Evans, and Hajishirzi]{duan2024membership}
Michael Duan, Anshuman Suri, Niloofar Mireshghallah, Sewon Min, Weijia Shi, Luke Zettlemoyer, Yulia Tsvetkov, Yejin Choi, David Evans, and Hannaneh Hajishirzi.
\newblock Do membership inference attacks work on large language models?
\newblock arXiv:2402.07841, 2024.

\bibitem[Dubi{\'n}ski et~al.(2024)Dubi{\'n}ski, Kowalczuk, Boenisch, and Dziedzic]{dubinski2024cdi}
Jan Dubi{\'n}ski, Antoni Kowalczuk, Franziska Boenisch, and Adam Dziedzic.
\newblock {CDI}: {C}opyrighted data identification in diffusion models.
\newblock arXiv:2411.12858, 2024.

\bibitem[Epstein et~al.(2023)Epstein, Hertzmann, of~Human~Creativity, Akten, Farid, Fjeld, Frank, Groh, Herman, Leach, et~al.]{epstein2023art}
Ziv Epstein, Aaron Hertzmann, Investigators of Human~Creativity, Memo Akten, Hany Farid, Jessica Fjeld, Morgan~R Frank, Matthew Groh, Laura Herman, Neil Leach, et~al.
\newblock Art and the science of generative {AI}.
\newblock \emph{Science}, 380\penalty0 (6650):\penalty0 1110--1111, 2023.

\bibitem[Esser et~al.(2024)Esser, Kulal, Blattmann, Entezari, M{\"u}ller, Saini, Levi, Lorenz, Sauer, Boesel, et~al.]{esser2024scaling}
Patrick Esser, Sumith Kulal, Andreas Blattmann, Rahim Entezari, Jonas M{\"u}ller, Harry Saini, Yam Levi, Dominik Lorenz, Axel Sauer, Frederic Boesel, et~al.
\newblock Scaling rectified flow transformers for high-resolution image synthesis.
\newblock In \emph{41st International Conference on Machine Learning}, 2024.

\bibitem[Fu et~al.(2024)Fu, Tamir, Sundaram, Chai, Zhang, Dekel, and Isola]{fu2024dreamsim}
Stephanie Fu, Netanel Tamir, Shobhita Sundaram, Lucy Chai, Richard Zhang, Tali Dekel, and Phillip Isola.
\newblock {DreamSim}: {L}earning new dimensions of human visual similarity using synthetic data.
\newblock \emph{Advances in Neural Information Processing Systems}, 36, 2024.

\bibitem[Hayes et~al.(2017)Hayes, Melis, Danezis, and De~Cristofaro]{hayes2017logan}
Jamie Hayes, Luca Melis, George Danezis, and Emiliano De~Cristofaro.
\newblock {LOGAN}: {M}embership inference attacks against generative models.
\newblock arXiv:1705.07663, 2017.

\bibitem[Heath(2024)]{vergeExGoogle24}
Alex Heath.
\newblock Ex-{G}oogle {CEO} says successful {AI} startups can steal ip and hire lawyers to ‘clean up the mess’.
\newblock The Verge, 2024.

\bibitem[Hilprecht et~al.(2019)Hilprecht, H{\"a}rterich, and Bernau]{hilprecht2019monte}
Benjamin Hilprecht, Martin H{\"a}rterich, and Daniel Bernau.
\newblock Monte {C}arlo and reconstruction membership inference attacks against generative models.
\newblock \emph{Proceedings on Privacy Enhancing Technologies}, 2019.

\bibitem[Hu et~al.(2022)Hu, Salcic, Sun, Dobbie, Yu, and Zhang]{hu2022membership}
Hongsheng Hu, Zoran Salcic, Lichao Sun, Gillian Dobbie, Philip~S Yu, and Xuyun Zhang.
\newblock Membership inference attacks on machine learning: {A} survey.
\newblock \emph{ACM Computing Surveys}, 54\penalty0 (11s):\penalty0 1--37, 2022.

\bibitem[Li et~al.(2024)Li, Fu, Wang, Liu, Gao, Dai, and Han]{li2024unveiling}
Qiao Li, Xiaomeng Fu, Xi Wang, Jin Liu, Xingyu Gao, Jiao Dai, and Jizhong Han.
\newblock Unveiling structural memorization: {S}tructural membership inference attack for text-to-image diffusion models.
\newblock In \emph{ACM International Conference on Multimedia}, pages 10554--10562, 2024.

\bibitem[Mattern et~al.(2023)Mattern, Mireshghallah, Jin, Sch{\"o}lkopf, Sachan, and Berg-Kirkpatrick]{mattern2023membership}
Justus Mattern, Fatemehsadat Mireshghallah, Zhijing Jin, Bernhard Sch{\"o}lkopf, Mrinmaya Sachan, and Taylor Berg-Kirkpatrick.
\newblock Membership inference attacks against language models via neighbourhood comparison.
\newblock arXiv:2305.18462, 2023.

\bibitem[Melnik et~al.(2024)Melnik, Ljubljanac, Lu, Yan, Ren, and Ritter]{melnik2024video}
Andrew Melnik, Michal Ljubljanac, Cong Lu, Qi Yan, Weiming Ren, and Helge Ritter.
\newblock Video diffusion models: {A} survey.
\newblock arXiv:2405.03150, 2024.

\bibitem[Milmo(2024)]{guardian2024impossible}
Dan Milmo.
\newblock ‘impossible’ to create {AI} tools like {ChatGPT} without copyrighted material, {OpenAI} says.
\newblock The Guardian, 2024.

\bibitem[Minaee et~al.(2024)Minaee, Mikolov, Nikzad, Chenaghlu, Socher, Amatriain, and Gao]{minaee2024large}
Shervin Minaee, Tomas Mikolov, Narjes Nikzad, Meysam Chenaghlu, Richard Socher, Xavier Amatriain, and Jianfeng Gao.
\newblock Large language models: {A} survey.
\newblock arXiv:2402.06196, 2024.

\bibitem[Ramesh et~al.(2022)Ramesh, Dhariwal, Nichol, Chu, and Chen]{ramesh2022hierarchical}
Aditya Ramesh, Prafulla Dhariwal, Alex Nichol, Casey Chu, and Mark Chen.
\newblock Hierarchical text-conditional image generation with {CLIP} latents.
\newblock arXiv:2204.06125, 2022.

\bibitem[Raut and Singh(2024)]{raut2024generative}
Gaurav Raut and Apoorv Singh.
\newblock Generative {AI} in vision: A survey on models, metrics and applications.
\newblock arXiv:2402.16369, 2024.

\bibitem[Rombach et~al.(2022)Rombach, Blattmann, Lorenz, Esser, and Ommer]{rombach2022high}
Robin Rombach, Andreas Blattmann, Dominik Lorenz, Patrick Esser, and Bj\"orn Ommer.
\newblock High-resolution image synthesis with latent diffusion models.
\newblock In \emph{Proceedings of the IEEE/CVF Conference on Computer Vision and Pattern Recognition (CVPR)}, pages 10684--10695, 2022.

\bibitem[Samuelson(2023)]{samuelson2023ongoing}
Pamela Samuelson.
\newblock Ongoing lawsuits could affect everyone who uses generative {AI}.
\newblock \emph{Science}, 381:\penalty0 6654, 2023.

\bibitem[Schuhmann et~al.(2022)Schuhmann, Beaumont, Vencu, Gordon, Wightman, Cherti, Coombes, Katta, Mullis, Wortsman, et~al.]{schuhmann2022laion}
Christoph Schuhmann, Romain Beaumont, Richard Vencu, Cade Gordon, Ross Wightman, Mehdi Cherti, Theo Coombes, Aarush Katta, Clayton Mullis, Mitchell Wortsman, et~al.
\newblock Laion-{5B}: {A}n open large-scale dataset for training next generation image-text models.
\newblock \emph{Advances in Neural Information Processing Systems}, 35:\penalty0 25278--25294, 2022.

\bibitem[Shokri et~al.(2017)Shokri, Stronati, Song, and Shmatikov]{shokri2017membership}
Reza Shokri, Marco Stronati, Congzheng Song, and Vitaly Shmatikov.
\newblock Membership inference attacks against machine learning models.
\newblock In \emph{IEEE Symposium on Security and Privacy}, pages 3--18. IEEE, 2017.

\bibitem[Thompson(2024)]{thompson2024jokerai}
Stuart~A. Thompson.
\newblock We asked {A.I.} to create the {J}oker. {I}t generated a copyrighted image.
\newblock \emph{The New York Times}, 2024.

\bibitem[Whang et~al.(2023)Whang, Roh, Song, and Lee]{whang2023data}
Steven~Euijong Whang, Yuji Roh, Hwanjun Song, and Jae-Gil Lee.
\newblock Data collection and quality challenges in deep learning: {A} data-centric {AI} perspective.
\newblock \emph{The VLDB Journal}, 32\penalty0 (4):\penalty0 791--813, 2023.

\bibitem[Wu et~al.(2022)Wu, Yu, Li, Backes, and Zhang]{wu2022membership}
Yixin Wu, Ning Yu, Zheng Li, Michael Backes, and Yang Zhang.
\newblock Membership inference attacks against text-to-image generation models.
\newblock arXiv:2210.00968, 2022.

\bibitem[Zhai et~al.(2024)Zhai, Chen, Dong, Li, Shen, Gao, Su, and Liu]{zhai2024membership}
Shengfang Zhai, Huanran Chen, Yinpeng Dong, Jiajun Li, Qingni Shen, Yansong Gao, Hang Su, and Yang Liu.
\newblock Membership inference on text-to-image diffusion models via conditional likelihood discrepancy.
\newblock arXiv:2405.14800, 2024.

\bibitem[Zhang et~al.(2024)Zhang, Yu, Wen, Backes, and Zhang]{zhang2024generated}
Minxing Zhang, Ning Yu, Rui Wen, Michael Backes, and Yang Zhang.
\newblock Generated distributions are all you need for membership inference attacks against generative models.
\newblock In \emph{Proceedings of the IEEE/CVF Winter Conference on Applications of Computer Vision}, pages 4839--4849, 2024.

\end{thebibliography}
}

\end{document}